\documentclass[10pt,twocolumn,letterpaper]{article}

\usepackage{icb}
\usepackage{times}
\usepackage{epsfig}
\usepackage{graphicx}
\usepackage{amsmath}
\usepackage{amssymb}
\usepackage{booktabs}

\usepackage{url}
\usepackage{hyperref}
\usepackage{breakurl}
\usepackage{multirow}
\usepackage{array}
\usepackage{tabu}
\usepackage{xcolor}
\usepackage{algorithm}
\usepackage{algorithmic}
\usepackage{comment}

\usepackage[hang,flushmargin]{footmisc}

\icbfinalcopy 

\ificbfinal\fi
\begin{document}

\title{OCT Fingerprints: Resilience to Presentation Attacks}

\author{Tarang Chugh and Anil K. Jain\\
Department of Computer Science and Engineering\\
Michigan State University, East Lansing, Michigan 48824\\
{\tt\small \{chughtar, jain\}@cse.msu.edu}}

\maketitle
\thispagestyle{empty}

\begin{abstract}

Optical coherent tomography (OCT) fingerprint technology provides rich depth information, including internal fingerprint (papillary junction) and sweat (eccrine) glands, in addition to imaging any fake layers (presentation attacks) placed over finger skin. Unlike 2D surface fingerprint scans, additional depth information provided by the cross-sectional OCT depth profile scans are purported to thwart fingerprint presentation attacks. We develop and evaluate a presentation attack detector (PAD) based on deep convolutional neural network (CNN). Input data to CNN are local patches extracted from the cross-sectional OCT depth profile scans captured using THORLabs Telesto series spectral-domain fingerprint reader. The proposed approach achieves a TDR of $99.73\%$ @ FDR of $0.2\%$ on a database of $3,413$ bonafide and $357$ PA OCT scans, fabricated using 8 different PA materials. By employing a visualization technique, known as \textit{CNN-Fixations}, we are able to identify the regions in the OCT scan patches that are crucial for fingerprint PAD detection.
\end{abstract}

\section{Introduction}

\begin{figure}[t]
\centering
\includegraphics[trim=0.4cm 0cm 0.4cm 0.4cm, width=0.94\linewidth]{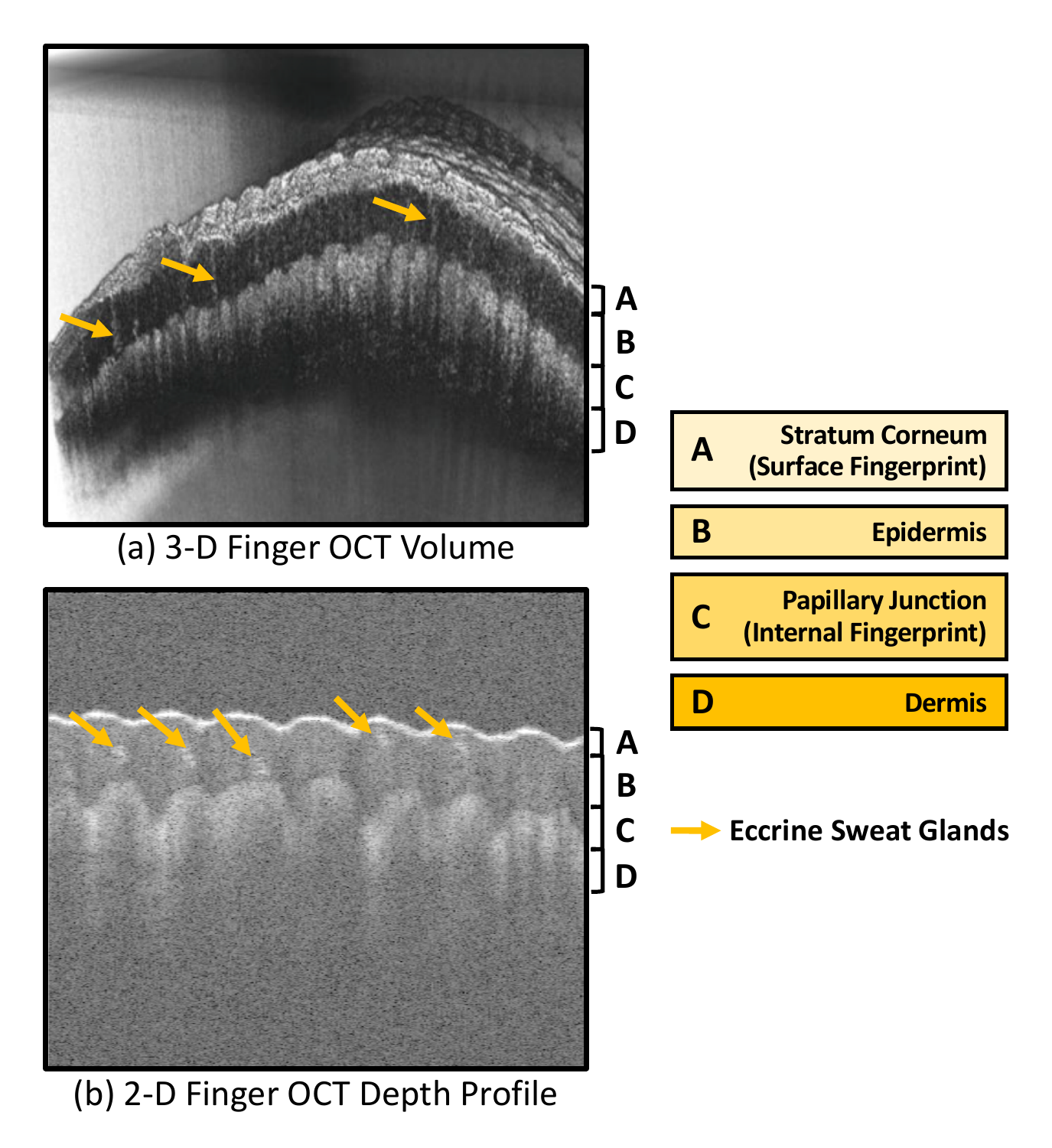}
\vspace{-2mm}
\caption{Different layers of a finger (stratum corneum, epidermis, papillary junction, and dermis) are distinctly visible in a OCT finger scan, along with helical shaped eccrine sweat glands in (a) 3-D finger OCT volume and (b) 2-D finger OCT depth profile. Note that (a) and (b) are OCT scans of different fingers. Image (a) is taken from~\cite{darlow2016performance}, and (b) is captured using THORLabs Telesto series (TEL1325LV2) SD-OCT scanner~\cite{scanner}.}
\label{fig:oct}
\vspace{-2.6mm}
\end{figure}

\begin{table*}[!htbp]
\caption{Existing studies on OCT-based fingerprint presentation attack detection.}
\label{tab:litrev}
\centering
\resizebox{\linewidth}{!}{
\begin{tabular}{ | p{2 cm} | >{\centering\arraybackslash}p{8 cm} | >{\centering\arraybackslash}p{3.1cm} | >{\centering\arraybackslash}p{3.5cm} | >{\centering\arraybackslash}p{3.5cm} | }
\toprule
\textbf{Study} & \textbf{Approach} & \textbf{OCT Technology} & \textbf{Database} & \textbf{Comments} \\ \midrule

 Cheng et al., 2006~\cite{cheng2006artificial} & 
 Averaged B-scan slices to generate 1D depth profile; performed auto-correlation analysis; B-scan is 2.2mm in depth and 2.4mm laterally & 
Imalux Corp. Time-domain OCT; capture time: 3s &
8 bonafide (8 fingers of one subject) and 10-20 impressions per PA, four PA materials &
Manual inspection of auto-correlation response \\ \midrule
 
Cheng et al., 2007~\cite{cheng2007vivo} & 
Extended~\cite{cheng2006artificial} by combining 100 B-Scans to create 3D representation; anisotropic resolution (4762 dpi, 254 dpi) &
Imalux Corp. Time-domain OCT; capture time: 300s for 100 scans &
One bonafide finger, one PA &
Visual analysis of 3D representation \\ \midrule
  
Bosen et al., 2010~\cite{bossen2010internal} &
Used fingerprint COTS for matching 3D OCT scans; scanned volume: 14mm x 14mm x 3mm; discussed detection of eccrine glands for PAD &
THORLabs Swept-source OCT (OCS1300SS); capture time: 20s for 3D volume &
153 impressions from 51 fingers for identification experiment; one PA material. &
Visual analysis for PAD; identification performance: FRR = 5\% @ FAR = 0.01\% \\ \midrule
   
Liu et al., 2010~\cite{liu2010biometric} &
Mapped subsurface eccrine glands with sweat pores on finger surface; exhibited repeatable matching of fingerprints based on sweat pores; discussed absence of sweat pores for fingerprint PAD&
Custom Spectral-domain OCT; capture time: 4min for 3D volume&
Nine bonafide impressions from three fingers, two PA materials &
Visual analysis of eccrine glands for PAD\\ \midrule
    
Nasiri-Avanaki et al., 2011~\cite{nasiri2011anti} &
Used a dynamic focus \textit{en-Face} OCT to detect any layer placed over finger skin; discussed Doppler OCT to detect blood flow and sweat production for liveness detection &
Custom \textit{en-Face} OCT; capture time is not reported.&
One bonafide finger, one PA &
Visual analysis of one bonafide finger and one sellotape PA\\ \midrule

Liu et al., 2013~\cite{liu2013capturing} &
Auto-correlation analysis between adjacent B-Scans to determine blood flow in micro-vascular pattern &
Swept-source OCT; capture time: 20s &
One bonafide with and w/o inhibited blood flow &
Exhibited repeatable signs of vitality \\ \midrule

Meissner et al., 2013~\cite{meissner2013defense} &
Detected number of helical eccrine gland ducts to distinguish bonafide vs PA, scanned volume: 4.5mm x 4mm x 2mm &
Swept-source OCT; capture time is not reported. &
Bonafide: $7,458$ images, cadavers: $330$ images, PA: $2,970$ images &
Manual PAD: $100\%$;  automated PAD: bonafide: $93\%$ and PA: $74\%$ success rate \\ \midrule

Darlow et al., 2016~\cite{darlow2016automated} &
Detected double bright peaks in depth profile for thin PAs and autocorrelation analysis for thick PAs; 2 different resolutions; scanned volume: 13mm x 13mm x 3mm (500dpi) and 15mm x 15mm x 3mm (867 dpi) &
THORLabs Swept-source OCT (OCS1300SS); capture time: 20s for 3D volume &
Bonafide: $540$ scans from 15 subjects, PA: 28 scans; one PA material + sellotape &
PAD accuracy: 100\%  \\ \midrule

Darlow et al., 2016~\cite{darlow2016damage} &
Measured ridge frequency consistency of the internal fingerprint in non-overlapping blocks; &
THORLabs Swept-source OCT (OCS1300SS) &
Bonafide: 20 scans, PA 20 scans; one PA material &
PAD accuracy: 100\%  \\ \midrule

Liu et al., 2019~\cite{liu2019high} &
Analyzed order and magnitude of bright peaks in 1-D depth signals to detect PAs with different thickness; scanned volume: 15mm x 15mm x 1.8mm&
Custom Spectral-domain OCT &
Bonafide: 30 scans from 15 subjects, PA: 60 scans; four PA materials &
Contact-based (glass platen) OCT scanner; PAD accuracy: 100\% \\ \midrule

\textbf{Proposed Approach} &
Trained a deep CNN model using overlapping patches extracted from detected finger depth profile in B-Scans; B-scan is 1.8mm in depth and 14mm laterally &
THORLabs Spectral-domain OCT (TEL1325LV2); capture time: $<1$s &
Bonafide: 3,413 scans from 415 subjects, PA: 357 scans, eight PA materials &
Five-fold cross-validation; TDR = 99.73\% @ FDR = 0.2\% \\ \bottomrule

\end{tabular}
}
\end{table*}


Fingerprint recognition based authentication systems have become ubiquitous with its footprint in a plethora of different applications such as mobile payments~\cite{applepay}, access control~\cite{swonger1980fingerprint}, international border crossing~\cite{obim} and national ID~\cite{aadhaar}. While the primary purpose of a fingerprint recognition system is to ensure a reliable and accurate user authentication, the security of the recognition system itself can be jeopardized by presentation attacks\footnote{The ISO standard \textit{IEC 30107-1:2016(E)}~\cite{isopad} defines presentation attacks as the \textit{``presentation to the biometric data capture subsystem with the goal of interfering with the operation of the biometric system".}}~\cite{marcel2014handbook,IARPAProject}. 

Most of the fingerprint recognition systems based on traditional readers (e.g., FTIR and capacitive technology) rely upon the friction ridge information  on the finger surface (\textit{i.e.} stratum corneum). This makes them highly prone to be fooled by presenting fabricated objects (presentation attack instruments or PAIs) with accurate imitation of another individual's bonafide\footnote{In the literature, the term \textit{live} fingerprint has been primarily used to refer a \textit{bonafide} fingerprint juxtaposed to spoof fingerprints. However, in the context of all forms of presentation attacks, bonafide fingerprint is a more appropriate term as some PAs such as fingerprint alterations also exhibit characteristics of liveness.} fingerprint ridge-valley structures~\cite{matsumoto2002impact}. Commonly available and inexpensive materials, such as play-doh, gelatin, wood glue, etc., have been utilized to fabricate fingerprint PAIs, capable of circumventing a fingerprint recognition system security with a reported success rate of more than $70\%$~\cite{biggio2012security, chugh2018fingerprint_icb19}. Such deception may permit unauthorized access to an impostor.

Given the increasing possibilities to realize presentation attacks (PAs), there is now an urgent requirement for robust PAD as a first line of defense to ensure the security of a fingerprint recognition system. In response to this, a series of fingerprint Liveness Detection (LivDet) competitions~\cite{ghiani2017review} have been held since 2009 to advance state-of-the-art and benchmark the proposed PAD solutions, with the latest competition held in 2019~\cite{orru2019livdet}. Generally, PAs can be detected by either: (i) software-based approaches, \textit{i.e.} extract features from the captured fingerprint image~\cite{mura2018livdet, marasco2015survey, marcel2014handbook}, or (ii) hardware-based approaches, i.e. using sensor(s) to gather evidence of the liveness of the subject~\cite{lapsley1998anti, rowe2006multispectral, hogan2019optical}.

Software-based solutions typically work with the grayscale surface fingerprint image (or a sequence of images) captured by a typical contact-based FTIR or capacitive fingerprint reader. The software-based approaches have explored (i) \textit{hand-crafted features}, such as anatomical~\cite{marcialis2010analysis}, physiological~\cite{marasco2012combining}, and texture-based~\cite{ghiani2013fingerprint, gragnaniello2015local}, and (ii) \textit{learned features} via CNN~\cite{nogueira2016fingerprint, pala2017deep, chugh2018fingerprint, gajawada2019universal}. In the case of hardware-based approaches, special types of sensors are proposed to detect the characteristics of vitality, such as blood flow~\cite{lapsley1998anti}, skin distortion~\cite{antonelli2006fake}, odor~\cite{baldisserra2006fake}, skin color using multi-view reader~\cite{engelsma2018raspireader}, sub-surface fingerprint imaging using multispectral reader~\cite{tolosana2018towards, rowe2006multispectral}, and imaging internal finger structure using optical coherence tomography (OCT)~\cite{hogan2019optical}. \\

\begin{figure}[t!]
\centering
\includegraphics[trim=0.8cm 0.8cm 0.8cm 0.5cm, width=\linewidth]{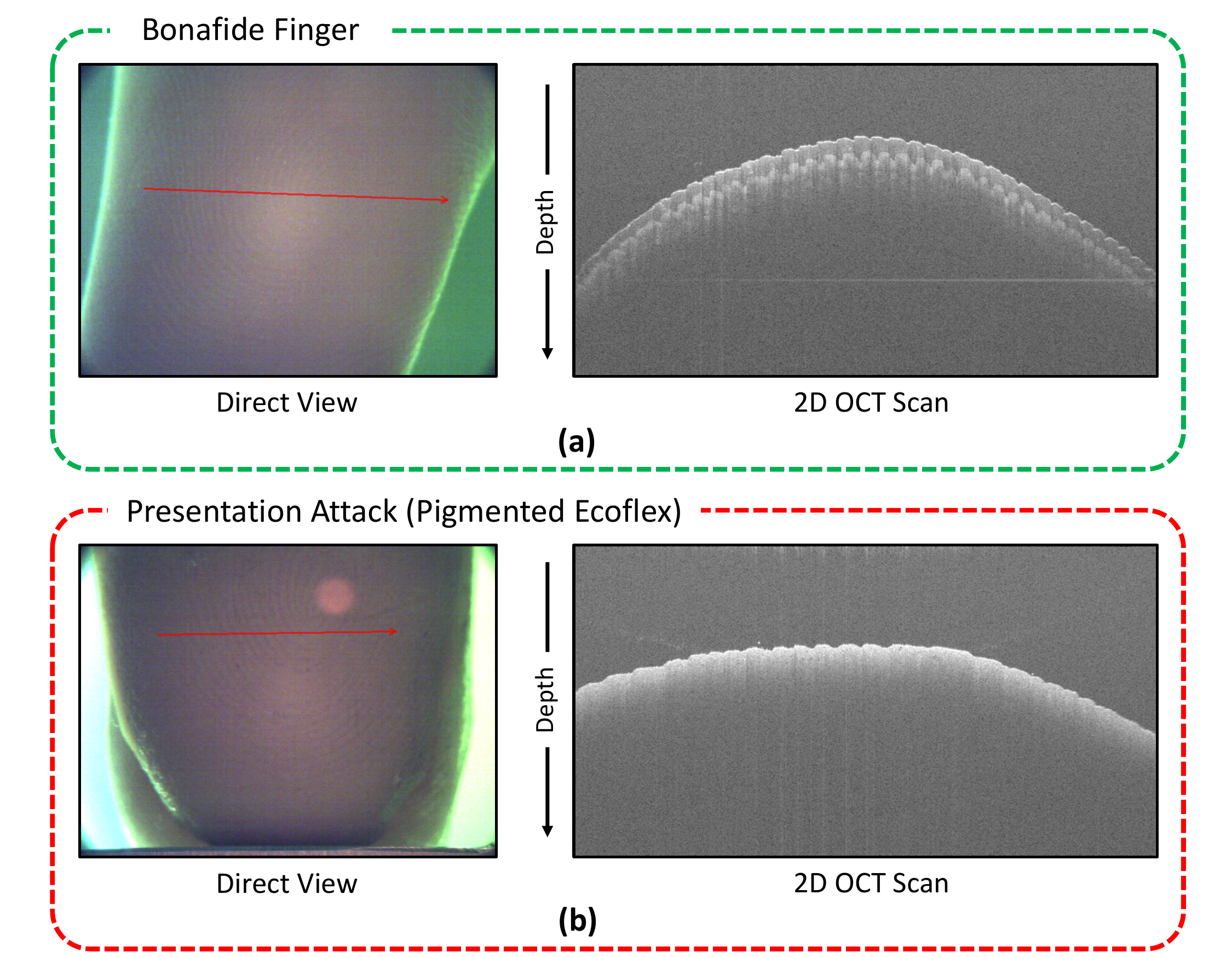}
\caption{Direct view images with red arrows presenting the scanned line and the corresponding cross-sectional B-scan for a (a) bonafide and a (b) pigmented ecoflex presentation attack.}
\label{fig:oct_scanline}
\end{figure}

Optical coherence tomography (OCT)~\cite{huang1991optical} technology allows non-invasive, high-resolution, cross-sectional imaging of internal tissue microstructures by measuring their optical reflections. Optical analogue to Ultrasound~\cite{wild1952application}, it utilizes low-coherence interferometry of near-infrared light ($900nm - 1325nm$) and is widely used in biomedical applications, such as ophthalmology~\cite{puliafito1995imaging}, oncology~\cite{hee1995quantitative}, dermatology~\cite{welzel2001optical} as well as applications in art conservation~\cite{liang2005face} and fingerprint presentation attack detection~\cite{moolla2019optical}. In an OCT scanner, a beam of light is split into a \textit{sample arm}, \textit{i.e.} a unit containing the object of interest, and a \textit{reference arm}, \textit{i.e.} a unit containing a mirror to reflect back light without any alteration. If the reflected light from the two arms are within coherence distance, it gives rise to an interference pattern representing the depth profile at a single point, also known as \textit{A-scan}. Laterally combining a series of A-scans along a line can provide a cross-sectional scan, also known as \textit{B-scan} (see Figs.~\ref{fig:oct} (b) and ~\ref{fig:oct_scanline}). Stacking multiple B-scans together can provide a 3D volumetric representation of the scanned object, or the object of our interest \textit{i.e.} internal structure of a finger (see Figure~\ref{fig:oct} (a)).

The human skin is a layered tissue with the outermost layer known as \textit{epidermis} and the external-facing sublayer of epidermis, where the friction ridge structure exists, is known as \textit{stratum corneum}. The layer below epidermis is known as \textit{dermis}, and the junction between epidermis and dermis layers is known as \textit{papillary junction}. The development of friction ridge patterns on papillary junction, which starts as early as in 10-12 weeks of gestation, results into the formation of surface fingerprint on stratum corneum~\cite{babler1991embryologic}. The surface friction ridge pattern scanned by traditional (optical and capacitive) fingerprint readers are merely an instance or a projection of the, so to say, a \textit{master print} existing on the papillary junction. There also exist helically shaped ducts in epidermis layer connecting the eccrine (sweat) glands in dermis to the sweat pores on surface. See Figure~\ref{fig:oct}. 

Since OCT enables imaging the 3D volumetric morphology of the skin tissue, including the subsurface fingerprint and other internal structures, it has great potential in detecting fingerprint presentation attacks. Existing fingerprint PAD studies in the literature have explored various OCT technologies such as time-domain, fourier-domain, and spectral domain, and developed hand crafted features to detect correlation between the skin layers, blood flow, eccrine glands, compute consistency in ridge frequency. These studies are summarized in Table~\ref{tab:litrev}. In this study, we utilize local patches ($150 \times 150$) extracted from the automatically segmented finger depth profile from input B-scan images to train a deep convolutional neural network. \\
The main contributions of this paper are:
\begin{enumerate}
\item Proposed a deep convolutional neural network based PAD approach trained on local patches containing finger depth profile from cross-sectional B-Scans.
\item Evaluated the proposed approach on a database of 3,413 bonafide and 357 PA OCT B-scans fabricated using 8 different PA materials and achieved a TDR of 99.73\% @ FDR of 0.2\% for PAD.
\item Identified the regions in the OCT scan patches that are crucial for fingerprint PAD detection by employing a visualization technique, known as \textit{CNN-Fixations}.
\end{enumerate}

\begin{figure*}[htbp]
\centering
\includegraphics[trim=0.2cm 3.1cm 0.2cm 3.1cm, width=0.94\linewidth]{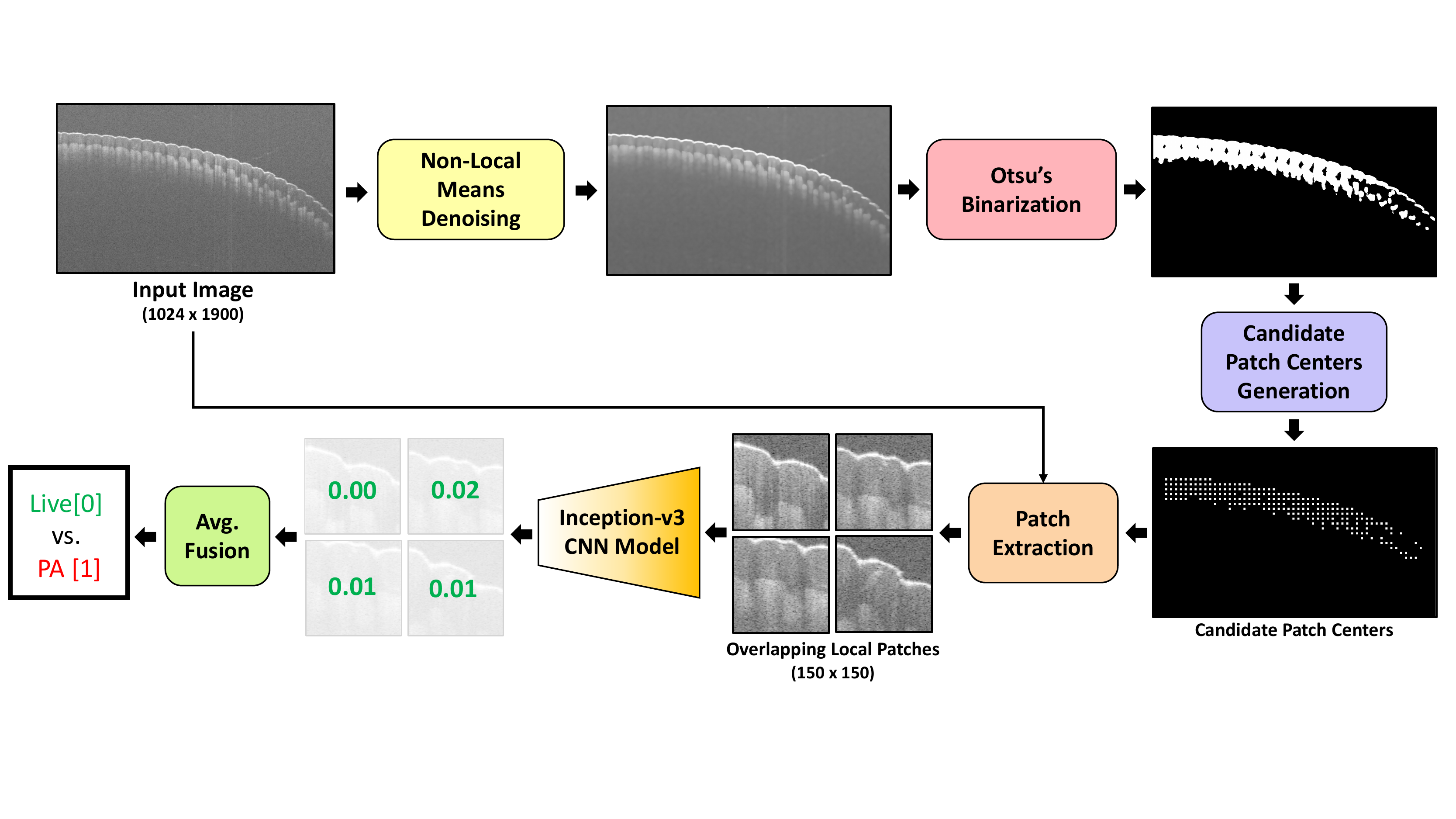}
\caption{An overview of the proposed fingerprint presentation attack detection approach utilizing local patches extracted from the segmented depth profiles from OCT B-scans.}
\label{fig:overview}
\end{figure*}

\section{Proposed Approach}
The proposed PAD approach includes two stages, an offline training stage and an online testing stage. The offline training stage involves (i) preprocessing the OCT images (noise removal and image enhancement), (ii) detecting region-of-interest (\textit{i.e.} finger depth profile), (iii) extracting local patches from the region-of-interest (ROI), (iv) and training CNN models on the extracted local patches. During the online testing stage, the final spoof detection decision is made based on the average of spoofness scores output from the CNN model for each of the extracted patches. An overview of the proposed approach is presented in Figure~\ref{fig:overview}.

\subsection{Preprocessing}
\begin{figure}[htbp]
\centering
\includegraphics[width=0.85\linewidth]{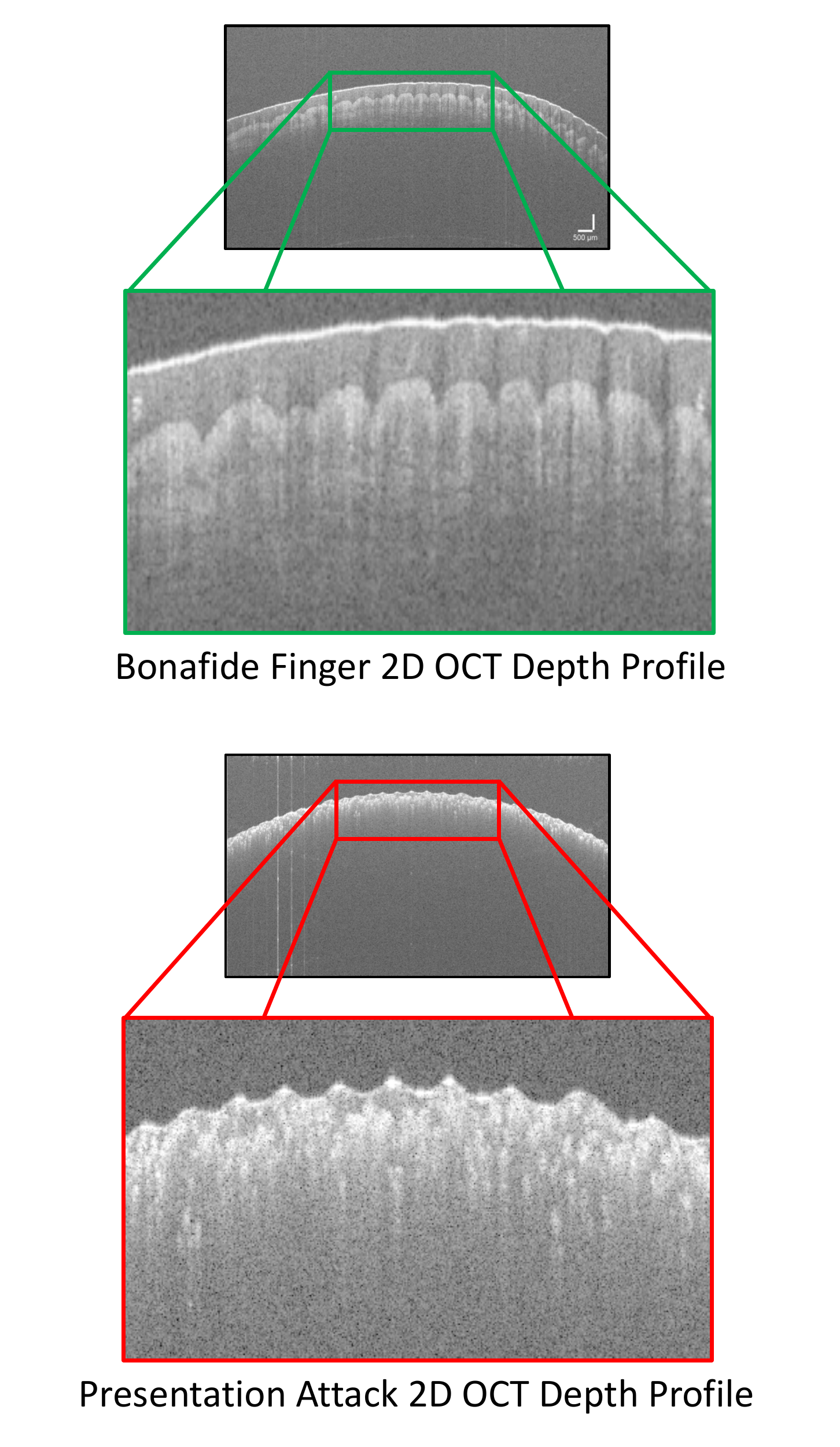}
\caption{Depth profile of a bonafide finger manifests a layered tissue anatomy quite distinguishable from the depth profile of a presentation attack without any specific structure.}
\label{fig:oct_diff}
\end{figure}

\begin{figure*}[htbp]
\centering
\includegraphics[width=\linewidth]{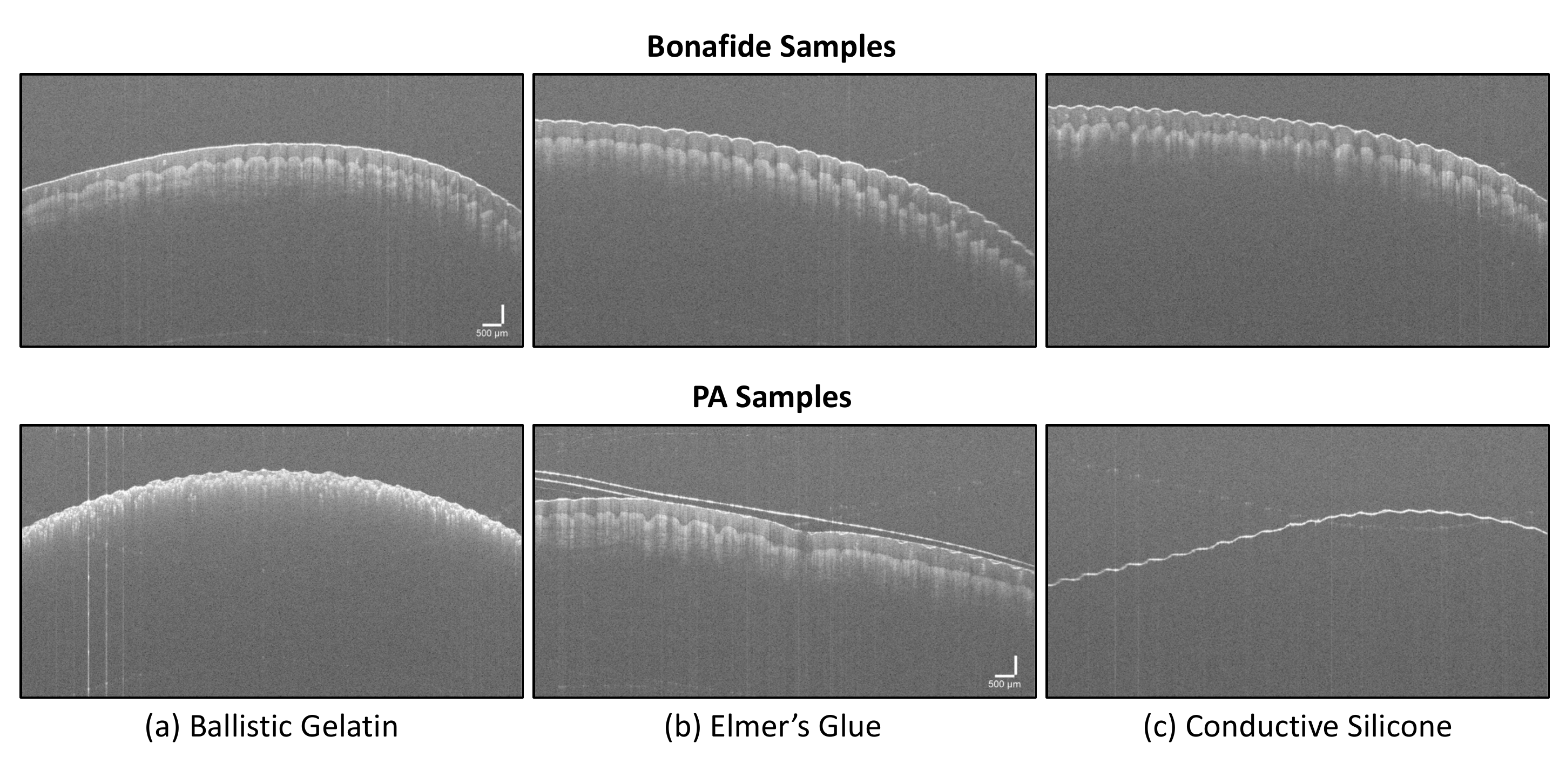}
\caption{Examples of bonafide and presentation attack samples from the OCT fingerprint database utilized in this study.}
\label{fig:samples}
\end{figure*}

Optical Coherent Tomography (OCT) 2D scans are grayscale images with height $=1024$ pixels and width $=1900$ pixels (see Figs.~\ref{fig:oct_diff} and ~\ref{fig:samples}). These images contain gaussian noise which makes the extraction of region-of-interest (finger depth profile) by simple thresholding prone to errors. We employ Non-Local Means denoising~\cite{buades2011non} that removes noise by replacing the intensity of a pixel with an average intensity of the similar pixels that may not be present close to each other (non-local) in the image. An optimized opencv python implementation\footnote{\url{https://opencv-python-tutroals.readthedocs.io/en/latest/py_tutorials/py_photo/py_non_local_means/py_non_local_means.html}} of Non-Local Means denoising, \textit{cv2.fastNlMeansDenoising()}, is used with \textit{filterStrength} = $20$, \textit{templateWindowSize} = $7$, and \textit{searchWindowSize} = $21$. After de-noising, a morphological operation of image dilation~\cite{forsyth2002computer}, with the kernel size of $5 \times 5$, is applied to enhance the image.

\subsection{Otsu's Binarization}
The characteristic differences between a bonafide and a presentation attack OCT image are primarily discernible in the finger depth profile region as shown in Figure~\ref{fig:oct_diff}. The pixel intensity histograms for the grayscale 2D OCT images are bimodal, with the first peak (high intensity values) referring to the finger depth profile region, while the second peak (low intensity values) referring to the background region. In order to segment out the finger depth profile, we apply Otsu's thresholding~\cite{otsu1979threshold} which finds an adaptive threshold, in the middle of the two peaks, to successfully binarize the input OCT images as shown in Figure~\ref{fig:overview}.

\subsection{Local Patch Extraction}
The binarized image generated after Otsu's binarization is raster scanned, with a stride of $30$ pixels (in both $x$ and $y$-axis), to identify the possible candidates for patch extraction. At each scanned pixel, a window of size $9 \times 9$ is evaluated and if more than $25\%$ of the pixels ($20$ out of $81$ pixels) in the window have non-zero values, the pixel is marked as a candidate for extracting a local patch. This rule is applied to guarantee sufficient depth information in the extracted patches. After the patch candidates are selected, a maximum of $60$ local patches of size $150 \times 150$ are extracted from the original image around the patch candidates. If there are more than $60$ candidates, the topmost candidates from each column (\textit{i.e.} the points closest to the surface fingerprint) are selected first, before moving to the next row. With the image width of $1900$ pixels and a stride of $30$ pixels, a maximum of 60 patches are sufficient to provide at least one pass of \textit{stratum corneum}. The patches are extracted such that the candidate is located at $(50, 75)$ in the $150 \times 150$ patch. This ensures that the extracted patches cover stratum corneum, epidermis, and papillary junction regions as shown in Figure~\ref{fig:overview}.

\begin{table}[t]
\centering
\caption{Summary of the OCT database used in this study.}
\label{tab:database}
\resizebox{\linewidth}{!}{
\begin{tabular}{ | >{\arraybackslash}p{7.1cm} | >{\raggedleft\arraybackslash}p{1.5cm} | } \hline
\textbf{Fingerprint Presentation Attack Material} & \textbf{\#Images} \\ \hline \hline
Ballistic Gelatin & $34$ \\ \hline
Clear Ecoflex & $7$ \\ \hline
Tan Ecoflex & $49$ \\ \hline
Yellow Pigmented Silicone & $57$ \\ \hline
Flesh Pigmented Ecoflex & $36$ \\ \hline
Nusil R-2631 Conductive Silicone & $128$ \\ \hline
Flesh Pigmented PDMS &$42$ \\ \hline
Elmer's Glue & $1$  \\ \hline
Bandaid & $3$ \\ \hline \hline
\textbf{Total PAs} & $\textbf{357}$ \\ \hline
\textbf{Total Bonafide} & $\textbf{3,413}$  \\ \hline
\end{tabular}
}
\end{table}

\subsection{Convolution Neural Networks}
With the success of AlexNet~\cite{krizhevsky2012imagenet} in ILSVRC-2012~\cite{russakovsky2015imagenet}, different deep CNN architectures have been proposed in literature, such as VGG, GoogleNet (Inception), Inception v2-v4, MobileNet, and ResNet. In this study, we utilize the Inception-v3~\cite{szegedy2016rethinking} architecture which has exhibited state-of-the-art performance in patch-based fingerprint presentation attack detection~\cite{chugh2017fingerprint, chugh2018fingerprint}. Our experimental results show that training the models from scratch, using local patches, performs better than fine-tuning a pre-trained network on image patches from other domains (e.g. FTIR fingerprint images). 

We utilized the TF-Slim library\footnote{\url{https://github.com/tensorflow/models/tree/master/research/slim}} implementation of the Inception-v3 architecture. The last layer of the architectures, a $1000$-unit softmax layer (originally designed to classify a query image into one of the the $1,000$ classes of ImageNet dataset) was replaced with a $2$-unit softmax layer for the two-class problem, i.e. Bonafide vs. PA. The output from the softmax layer is in the range $[0,1]$, defined as \textit{Spoofness Score}. The larger the spoofness score, the higher the likelihood that the input patch belongs to the PA class. For an input test image, the spoofness scores corresponding to each of the local patches, extracted from the input image, are averaged to give a \textit{Global Spoofness Score}. The optimizer used to train the network was RMSProp, with a batch size of 32, and an adaptive learning rate with exponential decay, starting at 0.01 and ending at 0.0001. Data augmentation techniques, such as random cropping, brightness adjustment, horizontal and vertical flipping, are employed to ensure the trained model is robust to the possible variations in fingerprint images.

\begin{figure}[htbp!]
\centering
\includegraphics[trim=0cm 0cm 0cm 0cm, width=\linewidth]{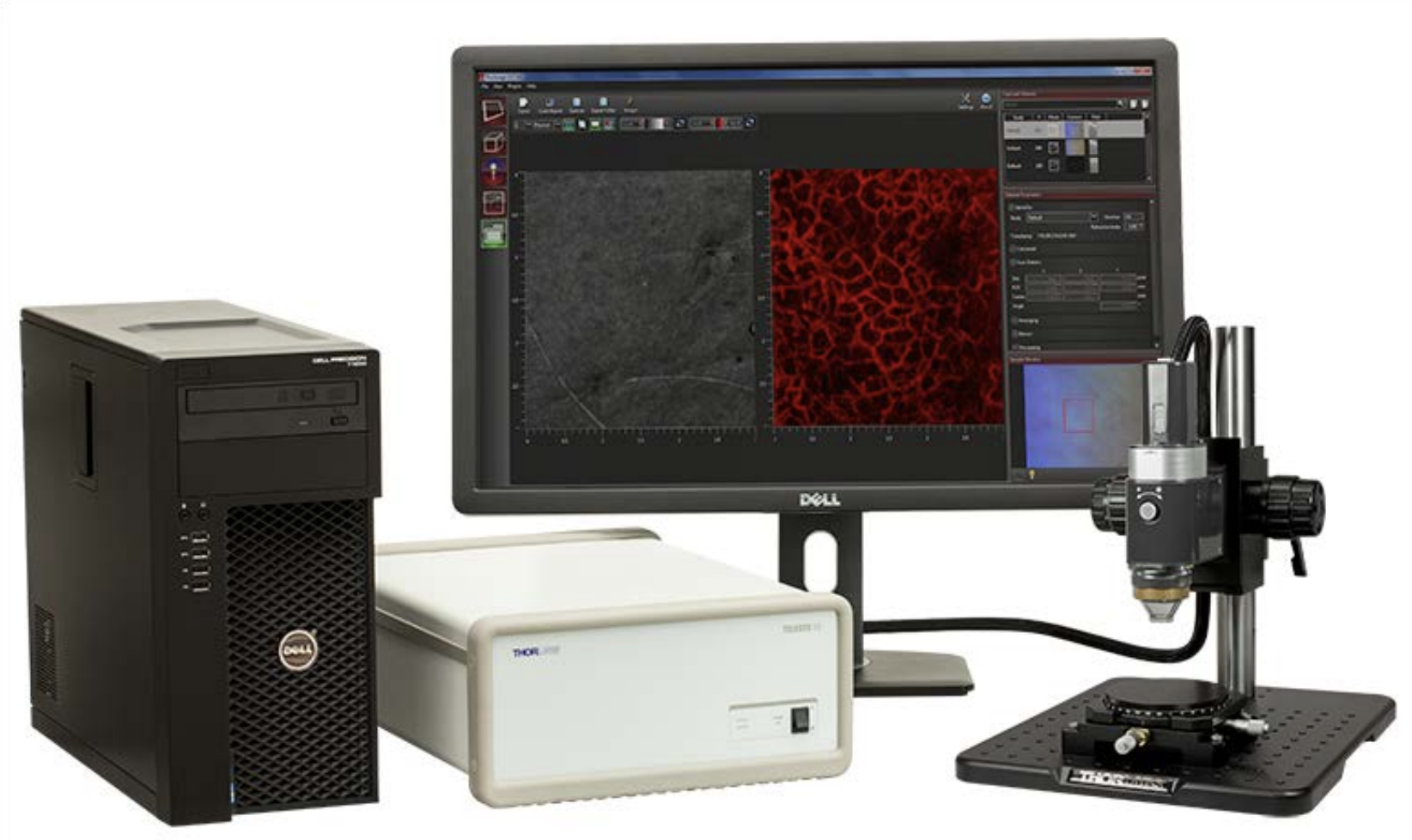}
\caption{Setup of a THORLabs Telesto series Spectral-domain OCT scanner (TEL1325LV2). Image taken from~\cite{scanner}.}
\label{fig:scanner}
\end{figure}

\section{Experimental Results}

\subsection{Presentation Attack OCT Database}
\label{sec:database}

A database of $3,413$ bonafide and $357$ presentation attack (PA) 2D OCT scans is utilized in this study. These scans are captured using THORLabs Telesto series (TEL1325LV2) Spectral-domain OCT scanner~\cite{scanner} (see Figure~\ref{fig:scanner}). Table~\ref{tab:database} lists the eight PA materials and the corresponding number of scans for each material type. Figure~\ref{fig:samples} presents few samples of bonafide and PA scans from this database. This dataset is collected at John Hopkins University Applied Physics Lab\footnote{\url{https://www.jhuapl.edu/}} as part of a large-scale evaluation under IARPA ODIN Project~\cite{IARPAProject} on Presentation Attack Detection.

\begin{table}[t]
\centering
\caption{Summary of the five-fold cross-validation and the performance achieved using Inception-v3 model.}
\label{tab:results}
\resizebox{\linewidth}{!}{
\begin{tabular}{ | >{\centering\arraybackslash}p{1.1cm} | >{\centering\arraybackslash}p{1.9cm} | >{\centering\arraybackslash}p{1.9cm} | >{\raggedleft\arraybackslash}p{4cm} | } \hline
 \multirow{2}{*}{\textbf{Fold}}  & \multicolumn{2}{c|}{\textbf{\# Images (Bonafide / PA)}} &  \multirow{2}{*}{\textbf{TDR (\%) @ FDR = 0.2\%}} \\ \cline{2-3}
& \textbf{Training} & \textbf{Testing} & \\ \hline \hline
I & (2,730 / 281) & (683 / 76) & $100.00$ \\ \hline
II & (2,730 / 283) & (683 / 74) & $98.63$ \\ \hline
III & (2730 / 288) & (683 / 71) & $100.00$ \\ \hline
IV & (2731 / 289) & (682 / 70)& $100.00$ \\ \hline
V & (2731 / 288) & (682 / 71) & $100.00$ \\ \hline \hline
\multicolumn{3}{|c|}{\textbf{Average}} & \textbf{99.73} (s.d. = 0.55) \\ \hline
\end{tabular}
}
\end{table}

\begin{figure}[htbp!]
\centering
\includegraphics[trim=0cm 0cm 0cm 0cm, width=\linewidth]{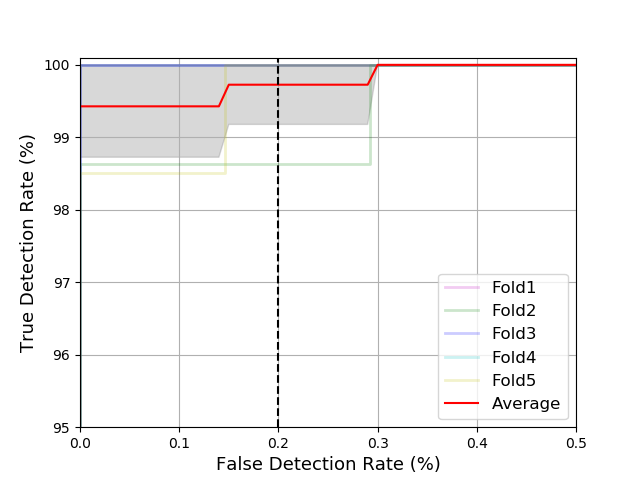}
\caption{ROC curves for the five-fold cross-validation experiments. The red curve represents the average performance with grayed region reflecting the confidence interval of one standard deviation.}
\label{fig:roc}
\end{figure}

\subsection{Results}
The proposed approach is evaluated using five-fold cross-validation. Table~\ref{tab:results} presents the training and testing set details for each fold\footnote{Note that all PA types are uniformly distributed among the five folds without repetition, therefore Elmer's Glue and Bandaid which have less than five samples are missing from some folds.}, along with the achieved PA True Detection Rate (\%) @ False Detection Rate $=0.2\%$. The selection of this metric is based on the requirements of IARPA ODIN program~\cite{IARPAProject} and represents the percentage of PAs able to breach the biometric system security when the reject rate of legitimate users $\le0.2\%$. Note that the proposed approach achieves an avg. TDR = $99.73\%$ (s.d. = $0.55$) @ FDR = $0.2\%$ for the five folds. Figure~\ref{fig:roc} presents the ROC curves for each of the five folds. In fold-II, only one bonafide scan was misclassified as PA due to incorrect segmentation.


\begin{figure*}[htbp!]
\centering
\includegraphics[trim=0cm 0cm 0cm 0cm, width=\linewidth]{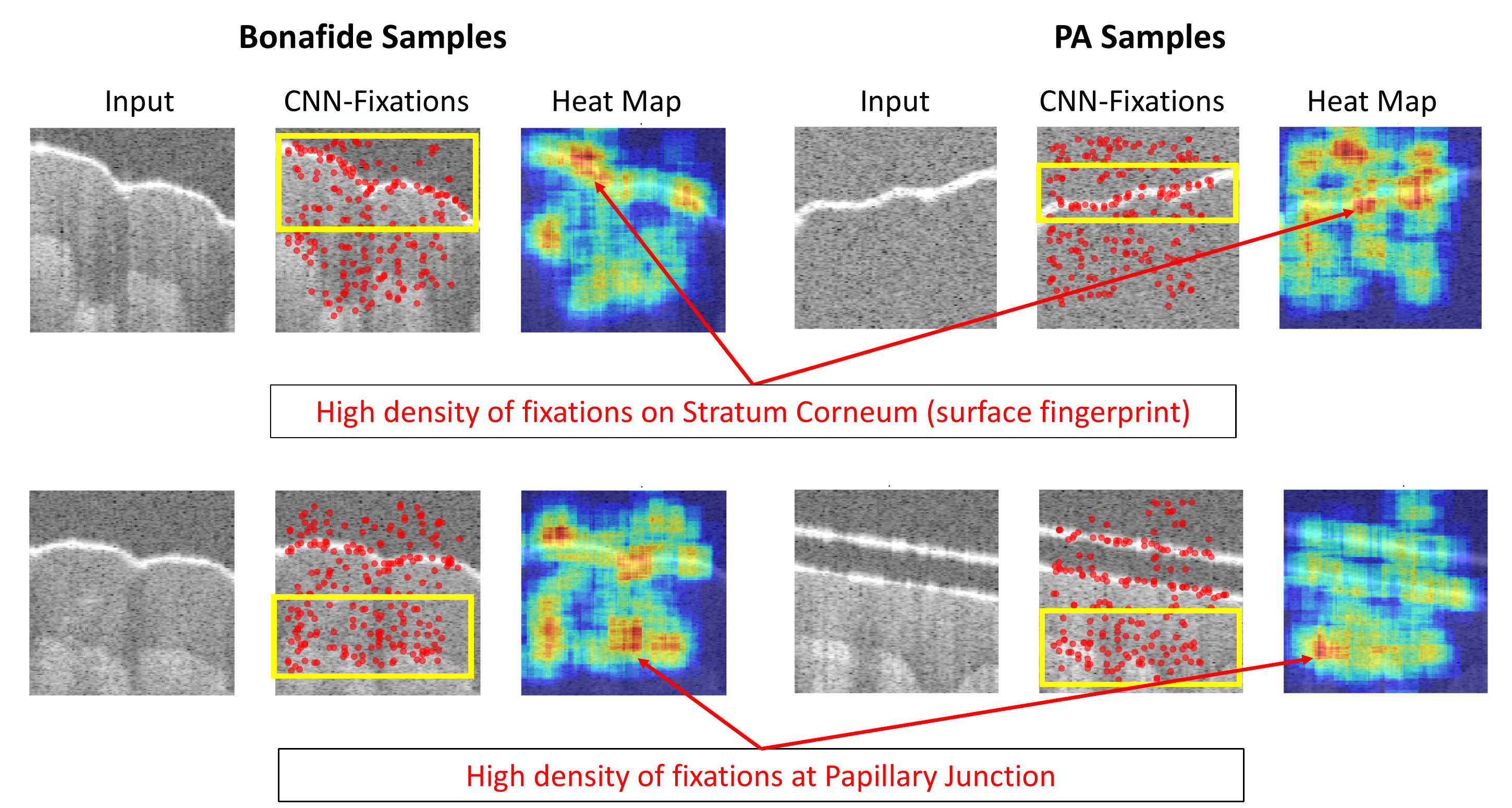}
\caption{Patches ($150 \times 150$) from bonafide and PA OCT B-scans input to the model are presented. The detected CNN-Fixations and a heat map presenting the density of CNN-Fixations are also shown. A high density of fixations are observed along the stratum corneum (surface fingerprint) and at papillary junction in both bonafide and PA patches.}
\label{fig:cnn_fixation_1}
\end{figure*}

\subsection{On understanding the learnings of CNN}

CNNs have revolutionized computer vision and machine learning research achieving unprecedented performance in many tasks. But these are usually treated as ``black boxes" shedding little light on their internal workings and without answering how they achieve high performance. One way to gain insights into what CNNs learn is through visual exploration, \textit{i.e.} to identify the image regions that are responsible for the final predictions. Towards this, visualization techniques~\cite{simonyan2013deep, selvaraju2017grad, mopuri2018cnn} have been proposed to supplement the class labels predicted by CNN, in our case bonafide or PA, with the discriminated image regions (or saliency maps) exhibiting class-specific patterns learned by CNN architectures. The visualization technique proposed in~\cite{mopuri2018cnn} exploits the learned feature dependencies between consecutive layers of a CNN to identify the discriminative pixels, called \textit{CNN-Fixations}, in the input image that are responsible for the predicted label. We utilize this visualization technique to understand the representation learning of our CNN models and identify the crucial regions in OCT images responsible for final predictions. Figs.~\ref{fig:cnn_fixation_1} presents CNN-Fixations and the corresponding density heatmaps for two bonafide and two PA image patches that are correctly classified. We observe that there is a high density of fixations along stratum corneum and at papillary junction, suggesting that these are definitely crucial regions in distinguishing bonafide vs PA OCT patches. Note that the only misclassified sample in Fold-II was due to incorrect segmentation, otherwise it would be useful to observe the CNN-Fixations that led to an incorrect prediction.

\section{Conclusions}
The penetrative power of optical coherent tomography (OCT) to image the internal tissue structure of human skin in a non-invasive manner presents a great potential to investigate robustness against fingerprint presentation attacks. We propose and demonstrate a learning-based approach to differentiate between bonafide (live) and eight different types of presentation attacks (spoofs). The proposed approach utilizes local patches automatically extracted from the finger depth profile in 2D OCT B-scans to train an Inception-v3 network model. Our experimental results achieve a TDR of $99.73\%$ @ FDR of $0.2\%$ on a database of $3,413$ bonafide and $357$ PA scans. The crucial regions in the input images for PAD learned by the CNN models, namely \textit{stratum corneum} and \textit{papillary junction}, are identified using a visualization technique. In future, we will evaluate the generalization ability of the proposed approach against novel materials that are not seen by the model during training.

\section{Acknowledgement}
This research is based upon work supported in part by the Office of the Director of National Intelligence (ODNI), Intelligence Advanced Research Projects Activity (IARPA), via IARPA R\&D Contract No. 2017 - 17020200004. The views and conclusions contained herein are those of the authors and should not be interpreted as necessarily representing the official policies, either expressed or implied, of ODNI, IARPA, or the U.S. Government. The U.S. Government is authorized to reproduce and distribute reprints for governmental purposes notwithstanding any copyright annotation therein.

{\small
\bibliographystyle{ieee}
\bibliography{egbib}
}

\end{document}